\documentclass[final]{cvpr}

\usepackage{times}
\usepackage{epsfig}
\usepackage{graphicx}
\usepackage{amsmath}
\usepackage{amssymb}

\usepackage{url}            % simple URL typesetting
\usepackage{booktabs}       % professional-quality tables
\usepackage{amsfonts}       % blackboard math symbols
\usepackage{nicefrac}       % compact symbols for 1/2, etc.
\usepackage{microtype}      % microtypography
\usepackage{tabularx}
\usepackage{multirow}
\usepackage{multicol}
\usepackage{array}
\newcommand{\PreserveBackslash}[1]{\let\temp=\\#1\let\\=\temp}
\newcolumntype{C}[1]{>{\PreserveBackslash\centering}p{#1}}
\newcolumntype{R}[1]{>{\PreserveBackslash\raggedleft}p{#1}}
\newcolumntype{L}[1]{>{\PreserveBackslash\raggedright}p{#1}}
\newcolumntype{?}{!{\vrule width 0.6pt}}
\DeclareMathOperator*{\argmin}{argmin}
\usepackage{dsfont}
\usepackage{pifont}% http://ctan.org/pkg/pifont
\newcommand{\cmark}{\ding{51}}%
\newcommand{\xmark}{\ding{55}}%

\usepackage{float}
\restylefloat{table}
\usepackage{placeins}

\usepackage[pagebackref=true,breaklinks=true,colorlinks,bookmarks=false]{hyperref}

\def\cvprPaperID{4761} % *** Enter the CVPR Paper ID here
\def\confYear{CVPR 2021}

\begin{document}

%%%%%%%%% TITLE
\title{Few-shot 3D Point Cloud Semantic Segmentation}

\author{Na Zhao \quad Tat-Seng Chua \quad Gim Hee Lee \\
	Department of Computer Science, National University of Singapore\\
	{\tt\small \{nazhao, chuats, gimhee.lee\}@comp.nus.edu.sg}
}

\maketitle

%%%%%%%%% ABSTRACT
\begin{abstract}
Many existing approaches for 3D point cloud semantic segmentation are fully supervised. These fully supervised approaches heavily rely on large amounts of labeled training data that are difficult to obtain and cannot segment new classes after training. To mitigate these limitations, we propose a novel attention-aware multi-prototype transductive few-shot point cloud semantic segmentation method to segment new classes given a few labeled examples. Specifically, each class is represented by multiple prototypes to model the complex data distribution of labeled points. Subsequently, we employ a transductive label propagation method to exploit the affinities between labeled multi-prototypes and unlabeled points, and among the unlabeled points. Furthermore, we design an attention-aware multi-level feature learning network to learn the discriminative features that capture the geometric dependencies and semantic correlations between points. Our proposed method shows significant and consistent improvements compared to baselines in different few-shot point cloud semantic segmentation settings (\ie 2/3-way 1/5-shot) on two benchmark datasets. Our code is available at \small \url{https://github.com/Na-Z/attMPTI}.
\end{abstract}

%%%%%%%%% BODY TEXT
\section{Introduction}
Point cloud semantic segmentation is a fundamental computer vision problem, which aims to estimate the category of each point in the 3D point cloud representation of a scene. The outcome of 3D semantic segmentation can benefit various real-world applications, including autonomous driving, robotics, and augmented/virtual reality. However, point cloud semantic segmentation is a challenging task due to the unstructured and unordered characteristics of point clouds. Recently, a number of fully supervised 3D semantic segmentation approaches \cite{hu2020randla,huang2018recurrent,landrieu2017large,li2018pointcnn,qi2017pointnet,wang2019dynamic,ye20183d,zhao2020ps} have been proposed and have achieved promising performance on several benchmark datasets \cite{armeni20163d,dai2017scannet}. 
Nonetheless, their success relies heavily on the availability of large amounts of labeled training data that are time-consuming and expensive to collect. 
Moreover, these approaches follow the \textit{closed set} assumption which states that the training and testing data are drawn from the same label space. 
However, the closed set assumption is not strictly adhered to the dynamic real world, where new classes can easily occur after training. 
As a result, these fully supervised approaches suffer from poor generalization to new classes with only few examples.

\begin{figure}[t]
	\centering
	\includegraphics[scale=0.38]{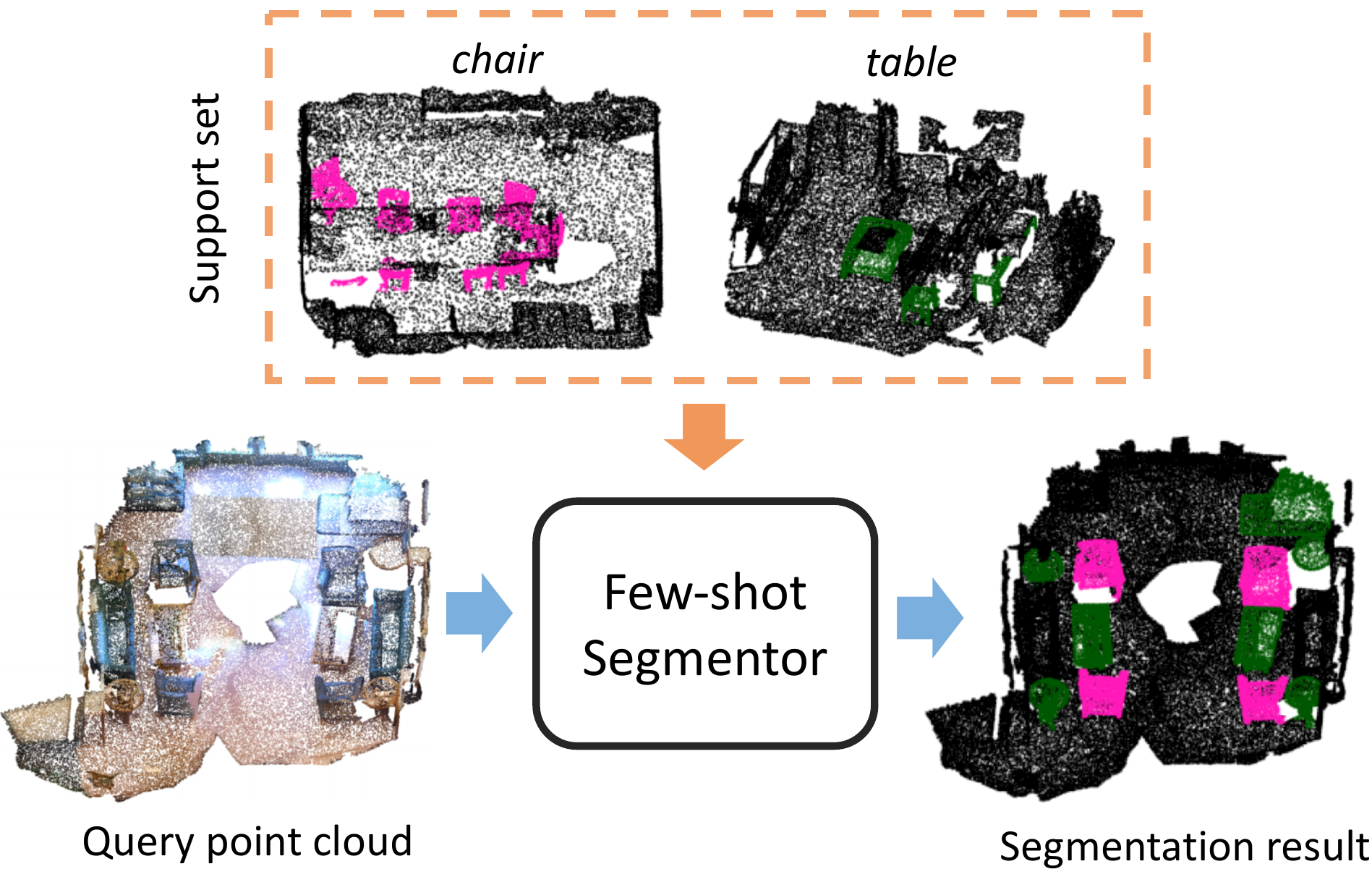}
	\caption{\small{Few-shot point cloud semantic segmentation task is to learn a segmentor that segments the query point cloud in terms of new classes with learned knowledge from the support examples. This figure illustrates an example with 2-way 1-shot setting.}}
	\label{fig:teaser}
	\vspace{-0.1in}
\end{figure}

Although several existing works used self-\cite{xie2020pointcontrast}, weakly- \cite{guinard2017weakly, xu2020weakly} and semi-supervised \cite{mei2019semantic} learning to mitigate the data hungry bottleneck in fully supervised 3D semantic segmentation, these approaches are still under the closed set assumption, where the generalization ability to new classes is overlooked.
The increasingly popular few-shot learning is a promising direction that allows the model to generalize to new classes with only a few examples.
In few-shot point cloud segmentation, our goal is to train a model to segment new classes given a few labeled point clouds, as illustrated in Figure \ref{fig:teaser}. 
We adopt the commonly used meta-learning strategy, \ie episodic training \cite{vinyals2016matching}, that learns over a distribution of similar few-shot tasks instead of only one target segmentation task. Each few-shot task consists of a few labeled samples (\textit{support set}) and unlabeled samples (\textit{query set}), and the model segments the query with learned knowledge from the support. 
Due to the consistency between the training few-shot task and the testing task, the model is 
endowed with better generalization ability that makes it less susceptible to overfitting to rare support samples. 
Despite the benefit of episodic training, few-shot point cloud segmentation still faces two major challenges on how to: 1) distill discriminative knowledge from scarce support that can represent the distributions of novel classes; and 2) leverage this knowledge to effectively perform segmentation.

In this paper, we propose a novel attention-aware multi-prototype transductive inference method for few-shot point cloud semantic segmentation. Specifically, our approach is able to model the complex distributions of the points within the point clouds of the support set, and perform the segmentation via transductive inference with the discriminative features extracted under the few-shot constraint.
We are motivated by the prototypical network \cite{snell2017prototypical}, which represents each class with a single prototype obtained from averaging the embeddings of labeled samples in the support.
We postulate that such uni-modal distribution assumption can be violated in point cloud segmentation due to the complex data distribution of points. In particular, the geometric structures of the points can vary largely within the same semantic class.  
Consequently, we propose to represent each class with multiple prototypes to better capture the complex distribution. 
Furthermore, it is important to learn discriminative features for the few-shot 3D point cloud semantic segmentation setting. To this end, we meticulously design an attention-aware multi-level feature learning network to learn the point-wise features by capturing the geometric dependencies and semantic correlations between the points. 
Subsequently, we perform the segmentation step in a transductive manner with the multiple prototypes in the learned feature space.
In contrast to the conventional prototypical network \cite{snell2017prototypical} that matches unlabeled instances with the class prototypes by computing their Euclidean distances, our transductive inference not only considers the relationships between the unlabeled query points and the multi-prototypes, but also exploits the relationships among the unlabeled query points.

The main contributions of this work are:
1) We are the first to study the promising few-shot 3D point cloud semantic segmentation task, which allows a model to segment new classes given a few or even one example(s).
2) We propose a novel attention-aware multi-prototype transductive inference method. 
Our designs of the attention-aware multi-level feature learning, and the affinity exploitation between multi-prototypes and unlabeled query points enable our model to obtain highly discriminative features and accomplish more precise segmentation in the few-shot scenario.
3) We conduct comprehensive experiments on the S3DIS and ScanNet datasets to demonstrate the superior performance of the proposed approach over baselines in different (\ie 2-/3-way 1-/5-shot) few-shot point cloud segmentation settings. Specifically, our method improves over the fine-tuning baseline
in the challenging 3-way 1-shot setting by 52\% and 53\% on the S3DIS and ScanNet dataset, respectively.

\section{Related Work}
\paragraph{3D Semantic Segmentation.}
Many deep learning based approaches \cite{hu2020randla,huang2018recurrent,landrieu2017large,li2018pointcnn,qi2017pointnet,wang2019dynamic,ye20183d,zhao2020ps} are proposed to tackle 3D semantic segmentation using full supervisions, \ie point-wise ground truths. PointNet~\cite{qi2017pointnet} is the first work that designs an end-to-end deep neural network to segment raw point clouds instead of their transformed representations, \eg voxel grids and multi-view images. Despite its simplicity and efficiency, PointNet overlooks the important local information embedded in the neighboring points. DGCNN \cite{wang2019dynamic} addresses this issue by designing the EdgeConv module that can capture local structures. In our work, we make use of DGCNN as the backbone of our feature extractor to extract local geometric features and semantic features. 
Although these fully supervised approaches achieved promising segmentation performance, their requirement for large amounts of training data precludes their use in many real-world scenarios where training data is costly or hard to acquire. 
Moreover, these approaches can only segment a set of pre-defined classes that are seen during training.
To alleviate these limitations, we explore the direction of few-shot learning for 3D semantic segmentation. This enables the model to segment new classes by seeing just a few labeled samples.

\vspace{-0.2in}
\paragraph{Few-shot Learning.}
The goal of few-shot learning is to develop a classifier that is able to generalize to new classes with very few examples (\eg one example for the one-shot case).  
To address this challenging few-shot learning, several meta-learning approaches \cite{finn2017model,garcia2018few,munkhdalai2017meta,ravi2017optimization,santoro2016meta,snell2017prototypical,vinyals2016matching} have proposed to learn transferable knowledge from a collection of learning tasks and made significant progress. 
In particular, metric-based method \cite{garcia2018few,snell2017prototypical,vinyals2016matching} is notable because of its effectiveness in directly inferring labels for unseen classes during inference. The key idea in metric-based method is to learn a good metric function which is able to produce a similarity embedding space representing the relationship between labeled and unlabeled samples.
\textit{Matching network} \cite{vinyals2016matching} and \textit{Prototypical Network} \cite{snell2017prototypical} are two representative metric-based methods.
Both methods utilize deep neural network to map the support and query sets into an embedding space, and then apply a non-parametric method to predict classes for the query based on the support. Specifically, matching network leverages the weighted nearest neighbor method that represents a class by all its support samples, while prototypical network leverages the prototypical method that represents a class by the mean of its support samples. 
These two non-parametric methods become two extreme ends of the spectrum of complicated to simple data distribution modeling when applied to few-shot point cloud semantic segmentation. This is because the support samples for one class in the point cloud counterpart can contain a large number of points.
In this paper, we represent each class in point clouds somewhere between the two extremes with multiple prototypes and perform segmentation in a transductive manner.

\begin{figure*}[t]
	\centering
	\includegraphics[scale=0.55]{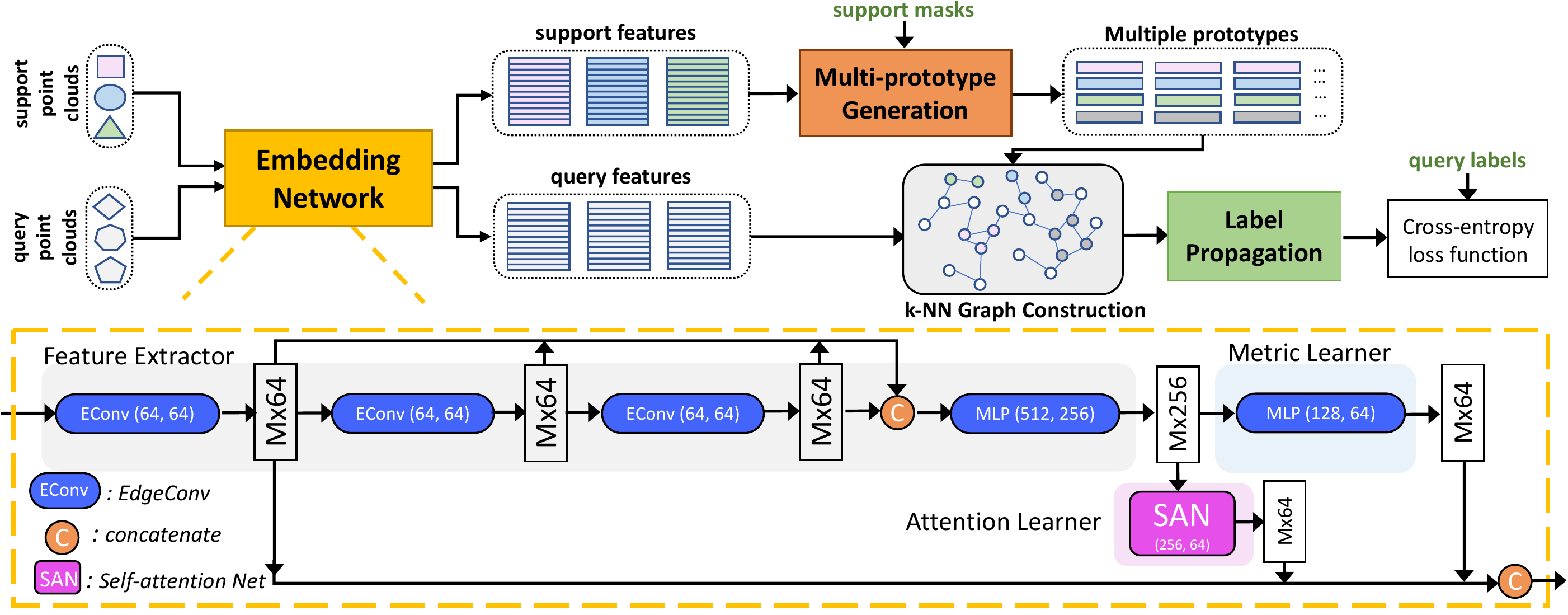}
	\caption{The architecture of our proposed method. This figure illustrates a 3-way 1-shot setting.}
	\label{fig:framework}
	\vspace{-0.1in}
\end{figure*}

\vspace{-0.2in}
\paragraph{Few-shot Image Segmentation.} 
All approaches mentioned previously focused on the few-shot image classification task. Only recently, several works \cite{dong2018few,liu2020crnet,nguyen2019feature, shaban2017one, wang2019panet, zhang2019pyramid, zhang2019canet} started to study few-shot learning on image segmentation by extending these meta-learning techniques to pixel levels. Most existing approaches \cite{dong2018few, nguyen2019feature, wang2019panet, zhang2019canet} leverage on metric-based techniques to solve a one-to-many matching problem between the support and query branch, where the support sample(s) of each class is represented as one global vector.
On the contrary, Zhang \etal \cite{zhang2019pyramid} considers the problem as many-to-many matching, where the support branch is represented as a graph with each element in the feature map of the support sample(s) as a node. 
%However, all these few-shot image segmentation approaches are designed for images and thus cannot directly applied or trivially adapted to the point cloud based counterpart due to huge data modality gap.
However, these few-shot image segmentation approaches learn image features by using convolution neural network (CNN) based architectures, which are not applicable to point cloud data due to the irregular structures of point clouds. Moreover, the properties of a good embedding space are different for point clouds (\cf Section \ref{sec:embedding_net}) and images.
In view of the differences, we design an attention-aware multi-level feature learning network and propose a novel attention-aware multi-prototype transductive inference method for the task of few-shot 3D point cloud semantic segmentation.

\section{Our Methodology}
\subsection{Problem Definition}
We align the training and testing of few-shot point cloud semantic segmentation with the episodic paradigm \cite{vinyals2016matching} that is commonly used in few-shot learning. Specifically, we train our model on a group of few-shot tasks sampled from a data set with respect to a training class set $C_{train}$, and then we test the trained model by evaluating it on another group of tasks sampled from a different data set with respect to new classes $C_{test}$, where $C_{test} \cap C_{train} = \varnothing$. Each few-shot task, a.k.a. an \textit{episode}, is instantiated as an $N$-way $K$-shot point cloud semantic segmentation task.
In each $N$-way $K$-shot episode, we are given a \textit{support} set, denoted as $S=\{(\textbf{P}_s^{1, k}, \textbf{M}^{1, k})_{k=1}^{K}, ...,  (\textbf{P}_s^{N, k}, \textbf{M}^{N, k})_{k=1}^{K}\}$, with $K$ labeled pairs of support point cloud $\textbf{P}_s^{n,k}$ and its corresponding binary mask $\textbf{M}^{n,k}$ for each of the $N$ unique classes.
Each point cloud $\textbf{P} \in \mathbb{R}^{M\times(3+f_0)}$ contains $M$ points associated with the coordinate information $\in \mathbb{R}^3$ and an additional feature $\in \mathbb{R}^{f_0}$, \eg color. We are also given a \textit{query} set, denoted as $Q=\{(\textbf{P}_q^{i}, \textbf{L}^{i})\}_{i=1}^{T}$, which contains $T$ pairs of query point cloud $\textbf{P}_q^{i}$ and its corresponding label $\textbf{L}^{i} \in \mathbb{R}^{M \times 1}$. Note that the ground-truth label $\textbf{L}$ is only available during training. The goal of $N$-way $K$-shot point cloud semantic segmentation is to learn a model $f_{\Phi}(\textbf{P}_q, S)$ that predicts the label distribution $\textbf{H} \in \mathbb{R}^{M \times (N+1)}$ for any \textit{query} point cloud $\textbf{P}_q$ based on $S$.
Formally, our training objective is to find the optimal parameters $\Phi^{*}$ of $f_{\Phi}(\textbf{P}_q, S)$ by computing:
\begin{equation}
	\Phi^{*} = \argmin_{\Phi} \mathbb{E}_{(S,Q) \sim \mathcal{T}_{train}} \Bigg[\sum_{(\textbf{P}_q^i, \textbf{L}^i) \in Q} J(\textbf{L}^i, f_{\Phi}(\textbf{P}_q^i, S)) \Bigg],
\end{equation}
where $\mathcal{T}_{train}$ denotes the training set containing all the episodes sampled from $C_{train}$, and $J(\cdot)$ is the loss function that will be defined in Section \ref{sec:loss_function}.

\subsection{Attention-aware Multi-prototype Transductive Inference Method}
Figure \ref{fig:framework} illustrates our attention-aware multi-prototype transductive inference framework. It consists of five components: 1) the \textbf{embedding network} that learns the discriminative features for the support and query point clouds; 2) the \textbf{multi-prototype generation} that produces multiple prototypes for each of the $N+1$ classes ($N$ semantic classes and one background class); 3) the \textbf{$k$-NN graph construction} that encodes both the cross-set (support-query) and intra-set (support-support, query-query) relationships within the embedding space; 4) the \textbf{label propagation} that diffuses labels through the whole graph along high density areas formed by the unlabeled query points; and 5) the \textbf{cross-entropy loss function} that computes the loss between the predicted labels and ground-truth labels of all the query points.

\subsubsection{Embedding Network} \label{sec:embedding_net}
\begin{figure}[t]
	\centering
	\includegraphics[scale=0.37]{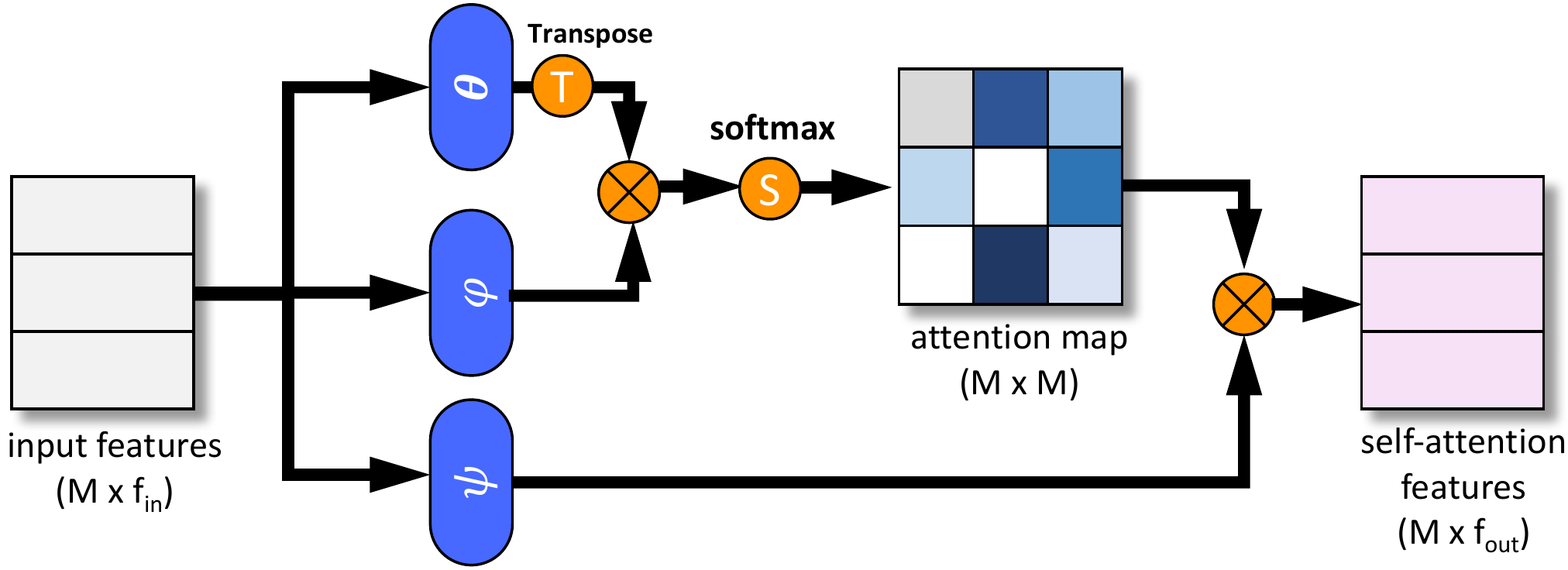}
	\caption{\small{The architecture of Self Attention Network (SAN). $\theta$, $\varphi$, and $\psi$ are linear embedding functions with trainable parameters.}}
	\label{fig:san}
	\vspace{-0.1in}
\end{figure}

The embedding network is the most important part of our model since both multi-prototype generation and $k$-NN graph construction are dependent on the learned embedding space. We expect this space to possess three properties: it can 1) encode the geometric structures of points based on local context; 2) encode the semantic information of points and their semantic correlation based on global context; and 3) quickly adapt to different few-shot tasks. To this end, we design an attention-aware multi-level feature learning network that incorporates three levels of features: local geometric features, global semantic features, and metric-adaptive features. Specifically, our embedding network is composed of three modules: \textit{feature extractor}, \textit{attention learner}, and \textit{metric learner}. 
We adopt DGCNN \cite{wang2019dynamic}, a dynamic graph CNN architecture, as the backbone of our feature extractor to respectively produce local geometric features (outputs of the first EdgeConv layer) and semantic features (outputs of the feature extractor). 
To further explore semantic correlation between points in the global context, we apply a self-attention network (SAN) on the generated semantic features. SAN allows the point-wise feature to aggregate the global contextual information of the corresponding point cloud in a flexible and adaptive manner. The architecture of SAN is illustrated in Figure \ref{fig:san}. 
In addition, we introduce the metric learner, \ie a stack of multi-layer perceptrons (MLP) layers to enable faster adaptability of the embedding space to different few-shot tasks since the feature extractor is updated with a slower learning rate (\cf the training details in Section \ref{sec:implementaion_details}).
The metric learner maps all point-wise features of support and query sets into a manifold space, where common distance functions (\eg euclidean distance or cosine distance) can be directly used to measure proximity between points. 
Finally, we concatenate the three levels of learned features together as the output of the embedding network.

\subsubsection{Multi-prototype Generation}
For each of the $N+1$ classes in the support set, we generate $n$\footnote{Although we can vary $n$ for different classes, we keep it uniform for simplicity.} prototypes to model the complex data distribution according to the few labeled samples in the episode. We cast the generation procedure as a clustering problem. While there can be different ways to cluster support points into multiple prototypes, we employ a simple strategy: sampling seed points and point-to-seed assignment based on the learned embedding space. Specifically, we sample a subset of $n$ seed points from a set of support points in one class using the farthest point sampling based on the embedding space. 
Intuitively, the farthest points in this space can inherently represent different perspectives of one class if the embedding space is learned well.
Let $\{\textbf{s}^c_i\}_{i=1}^n$ and $\{\textbf{f}^c_i\}_{i=1}^{m^c}$, where  $\{\textbf{s}^c_i\}_{i=1}^n \subset \{\textbf{f}^c_i\}_{i=1}^{m^c}$, denote the sampled seeds and all the $m^c$ support points belonging to the class $c$, respectively.
%$m^c >> n$. 
We compute the point-to-seed distance and take the index of the closest seed as the assignment of a point. Formally, the multi-prototypes $\boldsymbol{\mu}^c$ of class $c$ is given by:
\begin{align}
	\begin{split}
		& \boldsymbol{\mu}^c = \Big\{\boldsymbol{\mu}^c_1, ..., \boldsymbol{\mu}^c_n ~|~ \boldsymbol{\mu}^c_i=\frac{1}{|\mathcal{I}^{c*}_i|}\sum_{\textbf{f}^c_j \in \mathcal{I}^{c*}_i} \textbf{f}^c_j \Big\} \\
		&  \mathrm{s.t.} \quad \mathcal{I}^{c*} = \argmin_{\mathcal{I}^c} \sum_{i=1}^n \sum_{\mathbf{f}_j^c \in \mathcal{I}_i^c} \| \textbf{f}^c_j - \textbf{s}^c_i\|_2,
	\end{split}
\end{align}
where $\{\textbf{f}^c_i\}_{i=1}^{m^c}$ is partition into $n$ sets $\mathcal{I}^{c*} = \{\mathcal{I}^{c*}_1, ..., \mathcal{I}^{c*}_n\}$ such that $\mathbf{f}_j^c \in \mathcal{I}_i^{c*}$ is assigned to $\textbf{s}^c_i$.

\subsubsection{Transductive Inference}\label{sec:transductive_inference}
In addition to the similarity relations between each unlabeled query point and the labeled multi-prototypes, we also consider the similarity relations between pairs of unlabeled query points to exploit the ``smoothness" constraints\footnote{ Nearby points are most likely to have the same label.} among neighboring query points in our few-shot point cloud semantic segmentation task. 
To this end, we leverage on transductive inference to reason cross-set and intra-set relationships based on the embedding space. Concretely, we propose the use of transductive label propagation to construct a graph on the labeled multi-prototypes and the unlabeled query points, and then propagate the labels in the graph with random walk.

\vspace{-0.1in}
\paragraph{$k$-NN graph construction.} To mitigate the large number of query points, we construct a $k$ Nearest Neighbor (NN) graph instead of a fully-connected graph for computational efficiency.
Specifically, we take both the $n \times (N+1)$ multi-prototypes and $T \times M$ query points as nodes of a graph with size $V=n \times (N+1) + T \times M$. We construct a sparse affinity matrix, denoted as $\textbf{A} \in \mathbb{R}^{V\times V}$, by computing the Gaussian similarity between each node and its $k$ nearest neighbors in the embedding space:
\begin{equation}
	\textbf{A}_{ij} = \exp(-\frac{||\textbf{v}_i - \textbf{v}_j||^2_2}{2\sigma^2}),~ for~ \textbf{v}_j \in \mathcal{N}_k(\textbf{v}_i),
	\label{eq:affinitymatrix}
\end{equation}
where $\textbf{v}_i$ represents the node feature and $\sigma^2$ is the variance of the distance between two nodes.  
We follow \cite{iscen2019label} to let $\textbf{W} = \textbf{A} + \textbf{A}^\top$, this assures the adjacency matrix is non-negative and symmetric. 
Subsequently, we symmetrically normalize $\textbf{W}$ to yield $\textbf{S}=\textbf{D}^{-1/2}\textbf{W}\textbf{D}^{-1/2}$, where $\textbf{D}$ is the diagonal degree matrix with its diagonal value to be the sum of the corresponding row of $\textbf{W}$. 
Furthermore, we define a label matrix $\textbf{Y} \in \mathbb{R}^{V\times (N+1)}$, where the rows corresponding to labeled prototypes are one-hot ground truth labels and the rest are zero. 

\vspace{-0.1in}
\paragraph{Label propagation.}
Given $\textbf{S}$ and $\textbf{Y}$, label propagation iteratively diffuses labels through the graph according to:
\begin{equation}
	\textbf{Z}_{t+1} = \alpha \textbf{S} \textbf{Z}_t + (1-\alpha)\textbf{Y}.
\end{equation}
$\textbf{Z}_t \in \mathbb{R}^{V\times (N+1)}$ represents the predicted label distributions at iteration t, and $\alpha \in (0,1)$ is a parameter that controls the relative probability of the information from its adjacency nodes or its initial labels.
In \cite{zhou2004learning}, Zhou \etal show that sequence $\{\textbf{Z}_t\}$ converges to a closed-form solution:
\begin{equation}
	\textbf{Z}^* = (\textbf{I} - \alpha \textbf{S})^{-1}\textbf{Y}.
\end{equation}
We adopt the closed-form solution to directly compute the predictions $\textbf{Z}^*$ of label propagation.

\subsubsection{Loss Function} \label{sec:loss_function}
Once $\textbf{Z}^*$ is obtained, we first take the predictions corresponding to the $T$ query point clouds denoted as $\{\textbf{z}^i\}_{i=1}^T$, where $\textbf{z}^i \in \mathbb{R}^{M\times (N+1)}$ represents the predictions of the point cloud $\textbf{P}_q^i$. The prediction of each point in $\textbf{z}^i$ is then normalized into a probability distribution using the softmax function:
\begin{equation}
	\textbf{H}^i_{m,n}= \frac{\exp(\textbf{z}^i_{m,n})}{\sum_{j=1}^{N+1}\exp(\textbf{z}^i_{m,j})},
\end{equation}

Finally, we compute the \textit{cross-entropy loss} between $\{\textbf{H}^i\}_{i=1}^T$ and the ground truth labels $\{\textbf{L}^i\}_{i=1}^T$ as:
\begin{equation}
	J_{\Phi}= - \frac{1}{T}\frac{1}{M}\sum_{i=1}^{T}
	\sum_{m=1}^{M}\sum_{n=1}^{N+1} \mathds{1}[ \textbf{L}^i_m=n] \log(\textbf{H}^i_{m,n}),
\end{equation}
where $\Phi$ is the set of parameters of our model $f_{\Phi}(\textbf{P}_q, S)$. More precisely, $f_{\Phi}(\textbf{P}_q, S) = h(g_\Phi(\textbf{P}_q, \textbf{P}_s), \textbf{M})$ is a composite function of the embedding network $g_\Phi(.)$, and the multi-prototypes generation and transductive inference operations $h(.)$. It becomes apparent that the minimization of $J$ over the parameters $\Phi$ is governed by the affinity properties among the labeled multi-prototypes and unlabeled query points since the gradients have to flow through the parameter-less $h(.)$ into the embedding network $g_\Phi(.)$.

\section{Experiments}
We conduct experiments to evaluate the effectiveness of our method on two benchmark datasets. 
To the best of our knowledge, there is no prior study of few-shot point cloud semantic segmentation. Thus, we design the setup of the dataset, implementation details, and baselines for evaluation. 

\subsection{Datasets and Setup}
\paragraph{Datasets.} We evaluate on two datasets: 1) \textbf{S3DIS} \cite{armeni20163d} consists of 272 point clouds of rooms with various styles (\eg lobby, hallway, office, pantry). The annotation of the point clouds corresponds to 12 semantic classes plus one for the clutter. 
2) \textbf{ScanNet} \cite{dai2017scannet} consists of 1,513 point clouds of scans from 707 unique indoor scenes. The annotation of the point clouds corresponds to 20 semantic classes plus one for the unannotated space. 

\vspace{-0.2in}
\paragraph{Setup.} To customize the dataset to the few-shot learning setting, we evenly split the semantic classes in each dataset into two non-overlapping subsets based on the alphabetical order of the class names.
The splitting details are listed in Table \ref{tbl:data_split} of the supplementary material.
For each dataset, we perform cross-validation on the corresponding two subsets by selecting one split as the test class set $C_{test}$, while taking the remaining split as the training class set $C_{train}$. 

Since the number of points in the original rooms is large, we 
follow the data pre-processing strategy used in \cite{qi2017pointnet, wang2019dynamic} to divide the rooms into blocks using a non-overlapping sliding window of 1m$\times$1m on the $xy$ plane. 
As a result, S3DIS and ScanNet give 7,547 and 36,350 blocks, respectively. From each block, M $= 2,048$ points are randomly sampled. %as the input.

The training set $\mathcal{T}_{train}$ is constructed by including all the blocks that contain at least 100 points for any target class in $C_{train}$. During training, we randomly sample one episode from $\mathcal{T}_{train}$ using the following procedure: we first randomly choose $N$ classes from $C_{train}$, where $N < |C_{train}|$; and then a support set $S$ and a query set $Q$ are randomly sampled based on the chosen $N$ classes. The mask $\textbf{M}$ in the support set and the label $\textbf{L}$ in the query set are modified from the original point annotations accordingly to correspond to the chosen classes. 
The testing episodes are formed in a similar fashion, with the exception that we exhaustively iterate all the combinations of $N$ classes out of $C_{test}$ rather than randomly choosing $N$ classes. Specifically, we sample 100 episodes for each combination and use them as the $\mathcal{T}_{test}$ for evaluating each of the methods in our experiments. 
It is worth highlighting that the same point cloud can appear in both $\mathcal{T}_{train}$ and $\mathcal{T}_{test}$, but the annotations of this point cloud are different due to the difference in the classes of interest.

\begin{table*}[t]
	\centering
	\scalebox{0.84}{
		\begin{tabular}{L{1.6cm}?C{0.6cm} C{0.6cm} C{0.6cm}? C{0.6cm} C{0.6cm} C{0.6cm} ? C{0.6cm}  C{0.6cm}  C{0.6cm} ? C{0.6cm} C{0.6cm} C{0.6cm}}
			\hline\toprule[0.1pt]
			\multirow{3}{*}{\textbf{Method}} 
			& \multicolumn{6}{c?}{\textbf{2-way}}
			& \multicolumn{6}{c}{\textbf{3-way}} \\ \cline{2-13} 
			& \multicolumn{3}{c?}{\textbf{1-shot}} & \multicolumn{3}{c?}{\textbf{5-shot}}
			& \multicolumn{3}{c?}{\textbf{1-shot}} & \multicolumn{3}{c}{\textbf{5-shot}}
			\\ \cline{2-13} 
			& \multicolumn{1}{c}{S$^0$} & \multicolumn{1}{c}{S$^1$} &\multicolumn{1}{c?}{mean} 
			& \multicolumn{1}{c}{S$^0$} & \multicolumn{1}{c}{S$^1$} 
			& \multicolumn{1}{c?}{mean}              
			& \multicolumn{1}{c}{S$^0$} & \multicolumn{1}{c}{S$^1$} &\multicolumn{1}{c?}{mean} 
			& \multicolumn{1}{c}{S$^0$} & \multicolumn{1}{c}{S$^1$} 
			& \multicolumn{1}{c}{mean}  
			\\  \hline\toprule[0.1pt]
			FT  & 36.34 & 38.79  & 37.57  & 56.49  & 56.99  & 56.74  & 30.05  & 32.19 & 31.12  & 46.88  & 47.57  &  47.23 \\ \hline
			ProtoNet  & 48.39 & 49.98  &  49.19 &  57.34 & 63.22  & 60.28  & 40.81 & 45.07  &  42.94 & 49.05  & 53.42  & 51.24 \\ \hline
			AttProtoNet  & 50.98 & 51.90  &  51.44 & 61.02  & 65.25  & 63.14  & 42.16 & 46.76  &  44.46 & 52.20  & 56.20  & 54.20\\ \hline
			MPTI  & 52.27 &  51.48 &  51.88 & 58.93  & 60.56  & 59.75  & 44.27 & 46.92  &  45.60 & 51.74  & 48.57  & 50.16\\ \hline\toprule[0.1pt]
			\textbf{Ours}  & \textbf{53.77} & \textbf{55.94}  & \textbf{54.86}  & \textbf{61.67}  & \textbf{67.02}  & \textbf{64.35} & \textbf{45.18} & \textbf{49.27}  & \textbf{47.23}  &  \textbf{54.92} &  \textbf{56.79} & \textbf{55.86} \\ \hline\toprule[0.1pt]
		\end{tabular}
	}
	\caption{\small{Results on \textbf{S3DIS} dataset using mean-IoU metric (\%). S$^i$ denotes the split $i$ is used for testing.}}
	\label{tbl:exp-s3dis}
\end{table*}

\begin{table*}[t]
	\centering
	\scalebox{0.84}{
		\begin{tabular}{L{1.6cm}?C{0.6cm} C{0.6cm} C{0.6cm}? C{0.6cm} C{0.6cm} C{0.6cm} ? C{0.6cm}  C{0.6cm}  C{0.6cm} ? C{0.6cm} C{0.6cm} C{0.6cm}}
			\hline\toprule[0.1pt]
			\multirow{3}{*}{\textbf{Method}} 
			& \multicolumn{6}{c?}{\textbf{2-way}}
			& \multicolumn{6}{c}{\textbf{3-way}} \\ \cline{2-13} 
			& \multicolumn{3}{c?}{\textbf{1-shot}} & \multicolumn{3}{c?}{\textbf{5-shot}}
			& \multicolumn{3}{c?}{\textbf{1-shot}} & \multicolumn{3}{c}{\textbf{5-shot}}
			\\ \cline{2-13} 
			& \multicolumn{1}{c}{S$^0$} & \multicolumn{1}{c}{S$^1$} &\multicolumn{1}{c?}{mean} 
			& \multicolumn{1}{c}{S$^0$} & \multicolumn{1}{c}{S$^1$} 
			& \multicolumn{1}{c?}{mean}              
			& \multicolumn{1}{c}{S$^0$} & \multicolumn{1}{c}{S$^1$} &\multicolumn{1}{c?}{mean} 
			& \multicolumn{1}{c}{S$^0$} & \multicolumn{1}{c}{S$^1$} 
			& \multicolumn{1}{c}{mean}  
			\\  \hline\toprule[0.1pt]
			FT  & 31.55 & 28.94  & 30.25  &  42.71 & 37.24  & 39.98  & 23.99 & 19.10  & 21.55  & 34.93  &  28.10 & 31.52 \\ \hline
			ProtoNet  & 33.92 &  30.95 & 32.44  & 45.34  & 42.01  & 43.68  & 28.47 &  26.13 &  27.30 &   37.36 &  34.98 & 36.17 \\ \hline
			AttProtoNet  & 37.99 & 34.67  &  36.33 & 52.18  &  46.89 & 49.54  & 32.08 & 28.96  & 30.52 & 44.49  & 39.45  & 41.97 \\ \hline
			MPTI  & 39.27 & 36.14  & 37.71  &  46.90 & 43.59  & 45.25  & 29.96 & 27.26  & 28.61  & 38.14  & 34.36  & 36.25\\ \hline\toprule[0.1pt]
			\textbf{Ours}  & \textbf{42.55} & \textbf{40.83}  &  \textbf{41.69}  & \textbf{54.00}  & \textbf{50.32}  & \textbf{52.16}  & \textbf{35.23} & \textbf{30.72}  &  \textbf{32.98} & \textbf{46.74}  & \textbf{40.80}  & \textbf{43.77}\\ \hline\toprule[0.1pt]
		\end{tabular}
	}
	\caption{\small{Results on \textbf{ScanNet} dataset using mean-IoU metric (\%). S$^i$ denotes the split $i$ is used for testing.}}
	\label{tbl:exp-scannet}
	%\vspace{-0.1in}
\end{table*}

\vspace{-0.2in}
\paragraph{Evaluation metric.} For the evaluation metric, we adopt the widely used metric in point cloud semantic segmentation, \ie mean Interaction over Union (mean-IoU). In our few-shot setting, the mean-IoU is obtained by averaging over the set of testing classes $C_{test}$.

\subsection{Implementation Details}\label{sec:implementaion_details}
\paragraph{Framework details.} We illustrate the architecture and configuration of the embedding network in Figure \ref{fig:framework} (bottom). Following \cite{zhou2004learning}, the hyper-parameter $\alpha$ in label propagation is set to 0.99. The settings of the other three hyperparameters (\ie $n$ in multi-prototype generation, $k$ and $\sigma$ in the $k$-NN graph construction) are discussed in Section \ref{sec:results} and supplementary Section \ref{sec:hyperparams}.

\vspace{-0.2in}
\paragraph{Training.} We pre-train the feature extractor module on training set $\mathcal{T}_{train}$ by adding three MLP layers at the end of feature extractor as the segmentor over $C_{train}$. During pre-training, we set the batch size to 32 and train for 100 epochs. The pre-trained model is optimized by Adam with a learning rate of 0.001.
After initializing the feature extractor with the pre-trained weights, we use the Adam optimizer with an initial learning rate of 0.0001 for the feature extractor module, and an initial learning rate of 0.001 for the metric learner and attention learner modules, respectively. Both learning rates are decayed by half after 5,000 iterations. In each iteration, one episode is randomly sampled, and all the point clouds in the support and query set are augmented by Gaussian jittering and random rotation around z-axis.

\subsection{Baselines}
We design four baselines for comparison with our method.
\vspace{-0.4in}
\paragraph{Fine-tuning (FT).} We take the architecture of our pre-trained segmentation network as the backbone of this baseline. For fair comparison, we use the same pre-trained weights for this segmentation network and our method. Following the strategy in \cite{shaban2017one}, we fine-tune the trained segmentation network on samples from the support set and test on the query samples for each $N$-way $K$-shot task. To avoid overfitting, we only fine-tune the last three MLP layers.
\vspace{-0.2in}  
\paragraph{Prototypical Learning (ProtoNet).} We adapt the prototypical network \cite{garcia2018few} utilized in the few-shot image segmentation \cite{dong2018few, wang2019panet} task 
to few-shot point cloud segmentation. 
To extract the point-wise features for the support and query sets, we use similar architecture as our embedding network but replace SAN with a linear mapper that maps the features into the same dimension as SAN. Similarly, the feature extractor is initialized by the same pre-trained weights.
We represent each class by one prototype given by the mean feature of its support points. The predictions of query points are from its squared Euclidean distance with respect to the prototypes. 
%ProtoNet is trained with the cross-entropy loss, same as our method.
\vspace{-0.2in}
\paragraph{Attention-aware Prototypical Learning (AttProtoNet).} This baseline is an upgraded version of ProtoNet, where the self-attention mechanism is added into the embedding network. In other words, it uses the same architecture as our embedding network.
\vspace{-0.2in}
\paragraph{Multi-prototype Transductive Inference (MPTI).} This can be considered as a degraded version of our method, where the attention learner module (SAN) in the embedding network is replaced by a linear mapper similar to ProtoNet. 

\begin{figure*}[ht]
	\centering
	\includegraphics[scale=0.52]{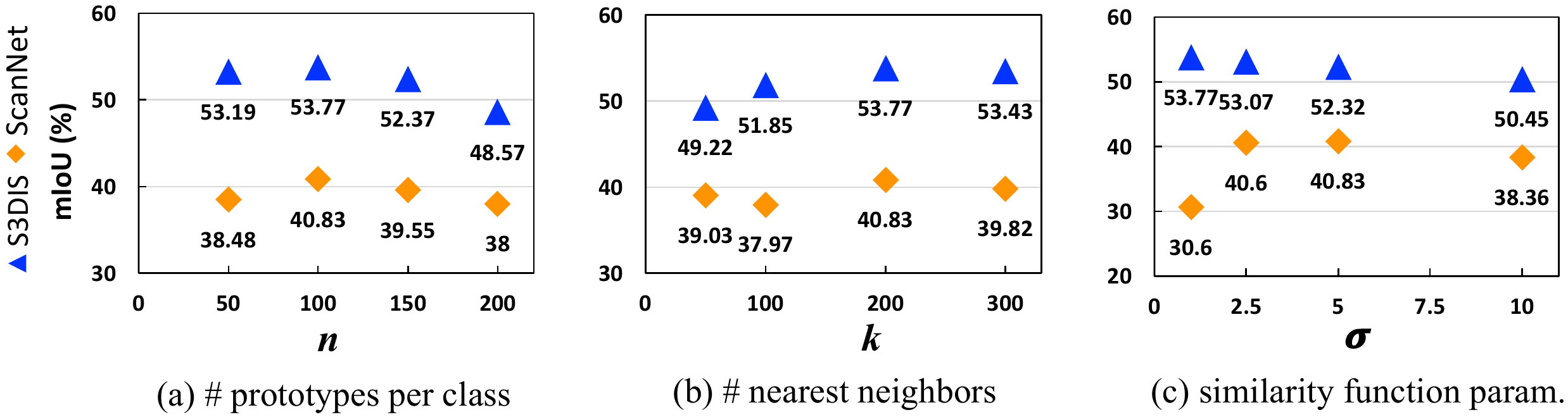}
	\caption{\small{Effects of three hyper-parameters under \textbf{2-way 1-shot} setting on S3DIS (S$^0$) and ScanNet (S$^1$) datasets.}}
	\label{fig:hyper-parameters}
	\vspace{-0.1in}
\end{figure*}

\subsection{Results and Analyses}\label{sec:results}
\paragraph{Comparison with baselines.} Table \ref{tbl:exp-s3dis} and \ref{tbl:exp-scannet} summarize the results of comparing our method to the baselines on S3DIS and ScanNet, respectively. It is not surprising that using more labeled samples, \ie larger $K$-shot leads to significant improvements for all the methods. We also observe that the performance of 3-way is generally lower than 2-way segmentation due to its higher difficulty. As can be seen from the two tables, our proposed method consistently and significantly outperforms the baselines in all four settings, \ie 2/3-way 1/5-shot on both datasets. 
Particularly, our method improves upon FT 
under the challenging 3-way 1-shot setting by around 52\% and 53\% on S3DIS and ScanNet dataset, respectively. Compared to ProtoNet, our method gains around 10\% and over 20\% improvements on S3DIS and ScanNet, respectively, when using only one sample, \ie one-shot. These improvements shows that our proposed method can obtain more useful knowledge from very limited data during inference. 
The superiority of our method as compared to AttProtoNet shows the contribution of our proposed multi-prototype transductive inference mechanism. Additionally, both the improvements of AttProtoNet over ProtoNet and the improvements of our method over MPTI demonstrate the capacity of self-attention network in exploiting semantic correlations between the points, which is very important in inferring point-wise semantic labels. 

An interesting observation is that the degraded version of our method, \ie MPTI clearly outperforms ProtoNet under the one-shot settings, but loses the gain under five shots. This is probably due to the naive multi-prototype generation of MPTI made it difficult to extract accurate multi-prototypes for a large number of support points if the embedding space is not learned well. This phenomenon also indicates the importance of incorporating the self-attention network, which helps in learning a more representative embedding space.

\vspace{-0.15in}
\paragraph{Ablation study of multi-level features.}
\begin{table}[t]
	\centering
	\scalebox{0.8}{
		\begin{tabular}{C{1.5cm} C{1.5cm} C{1.5cm} | C{1.5cm} ? C{1.5cm} }
			\hline \toprule[0.1pt]
			$f_{geometric}$ & $f_{semantic}$ & $f_{metric}$ & S3DIS & ScanNet \\ \hline 
			\cmark  & \xmark  & \xmark & 40.31  & 26.91    \\
			\xmark  & \cmark  & \xmark & 44.43 &   34.51 \\
			\xmark  & \xmark  & \cmark & 48.24  & 35.07  \\\hline		
			\cmark  & \cmark  & \xmark & 47.82  & 38.69 \\
			\cmark  & \xmark  & \cmark & 52.21  & 36.12   \\
			\xmark  & \cmark  & \cmark & 50.12  & 39.81  \\\hline
			\cmark  & \cmark  & \cmark & \textbf{53.77}  & \textbf{40.83} \\
			\hline \toprule[0.1pt]
	\end{tabular}}
	\caption{\small{Effects of different levels of features under \textbf{2-way 1-shot} setting on S3DIS (S$^0$) and ScanNet (S$^1$) datasets.}}
	\label{tab:ablation-features}
	\vspace{-0.1in}
\end{table}

We study the effects of various designs of the embedding network since it is one of the most important components of our method. We denote the levels of features, \ie local geometric feature, global semantic feature, and metric-adaptive feature as $f_{geometric}$, $f_{semantic}$, and $f_{metric}$, respectively. We select one or two level(s) of feature(s) as our embedded feature\footnote{Specifically, the output of the embedding network will be the selected feature or the concatenation of the two selected features.} for the estimation of its(their) contribution(s). The results of six variants are listed in Table \ref{tab:ablation-features}. From the perspective of individual feature, $f_{semantic}$ and $f_{metric}$ contribute more than $f_{geometric}$. This is reasonable since the embedding space are supposed to be semantic. By combining any two levels of features, we achieve varying improvements on the two datasets. Eventually, the integration of the three levels of features gives us the best performance on both datasets.

\vspace{-0.15in}
\paragraph{Effects of hyper-parameters.}
In Figure \ref{fig:hyper-parameters}, we illustrate the effects of three hyper-parameters (\ie, $n$, $k$, $\sigma$) for 2-way 1-shot point cloud semantic segmentation on one split of each dataset. 
As can be seen from Figure~\ref{fig:hyper-parameters}(a), increasing the number of prototypes per class $n$ can achieve better results, but overly large $n$ can lead to the over-fitting problem and cause adverse impact on the performance. 
As Figure~\ref{fig:hyper-parameters}(b) shows, there is a slight difference on performance between the two datasets when choosing a smaller $k$, \ie $k=50$. However, the overall trend is similar, and the selection of $k=200$ gives the best result on both datasets. As reported in \cite{liu2019learning, xiaojin2002learning},
$\sigma$ in the Gaussian similarity function used in the construction of the affinity matrix (see Eq.~\ref{eq:affinitymatrix}) plays a role in the performance of label propagation. We empirically find that $\sigma$ in different datasets has different optimal values. Specifically, $\sigma=1$ on S3DIS and $\sigma=5$ on ScanNet enable us to achieve the best performance, respectively.

\begin{figure}[t]
	\centering
	\includegraphics[scale=0.35]{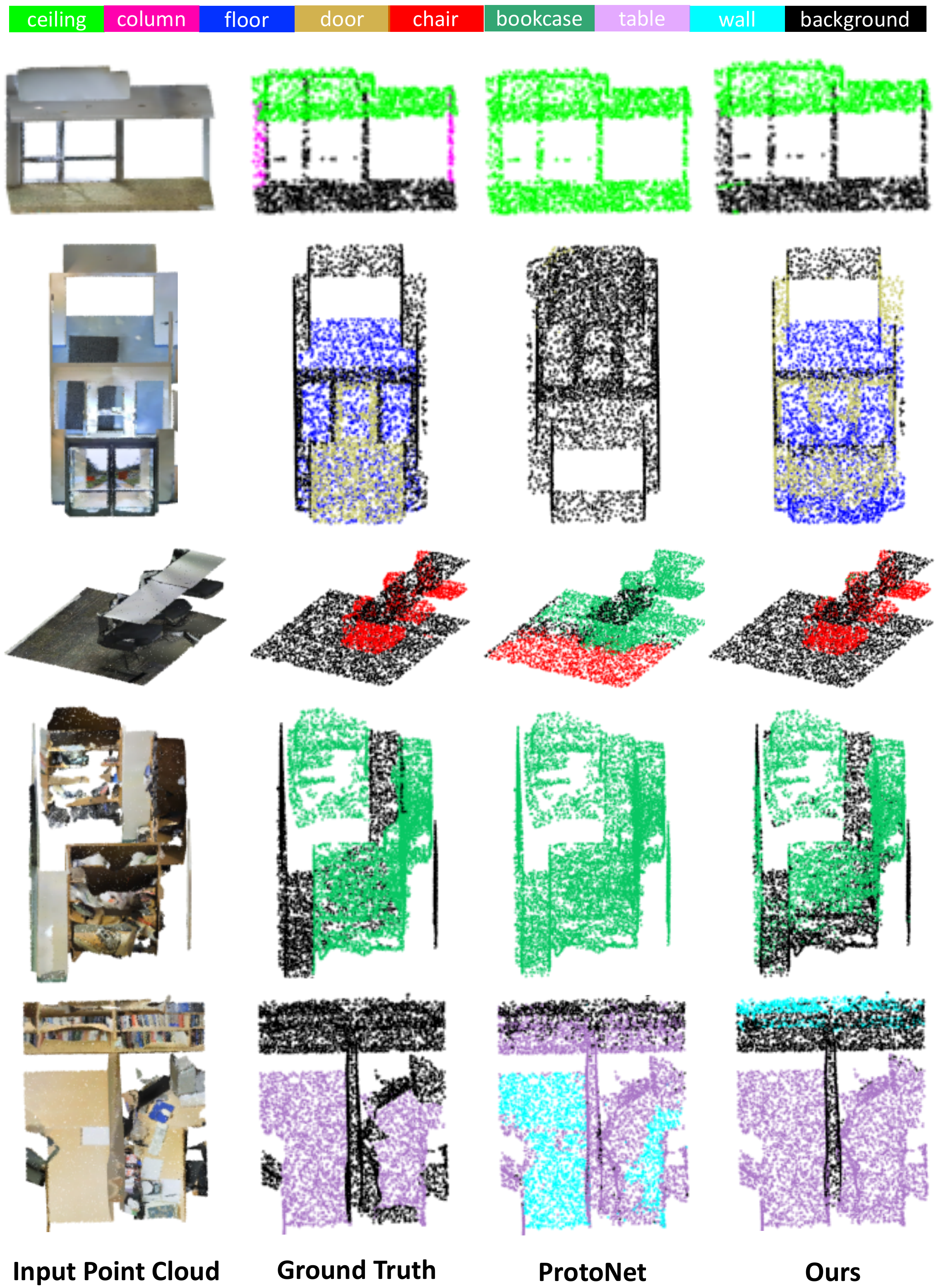} 
	\caption{Qualitative results of our method in\textit{ 2-way 1-shot} point cloud semantic segmentation on the \textbf{S3DIS} dataset in comparison to the ground truth and ProtoNet. Four combinations of  \textit{2-way} are illustrated from the top to bottom rows, \textit{i.e.}, ``\textit{ceiling, column}'' (first row), ``\textit{floor, door}'' (second row), ``\textit{chair, bookcase}'' (third and forth rows), ``\textit{table, wall}'' (last row).}
	\label{fig:s3dis-visualization}
\end{figure}

\subsection{Qualitative Results}
\begin{figure}[t]
	\centering
	\includegraphics[scale=0.34]{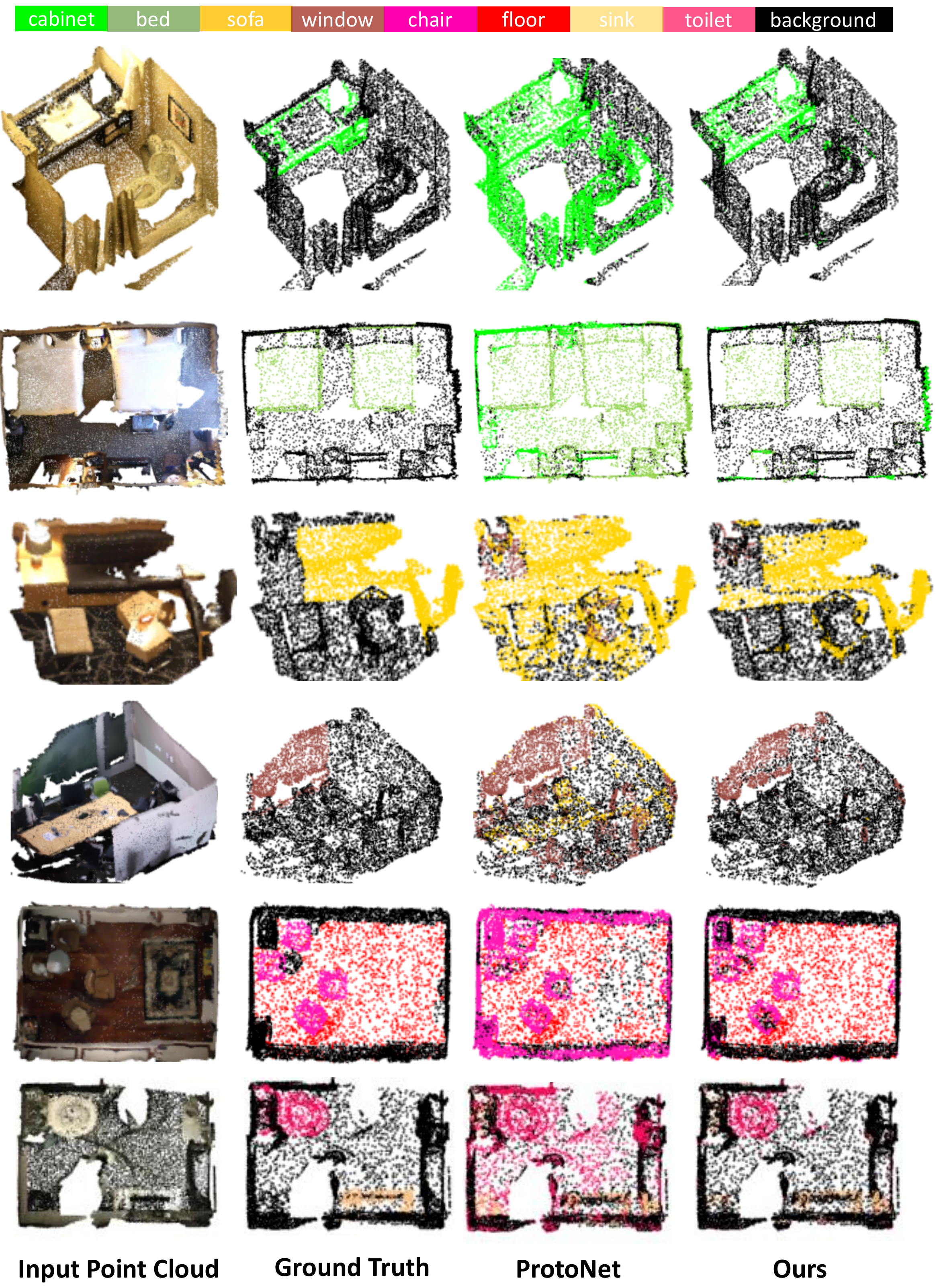} 
	\caption{Qualitative results of our method in \textit{2-way 1-shot} point cloud semantic segmentation on the \textbf{ScanNet} dataset in comparison to the ground truth and ProtoNet. Four combinations of  \textit{2-way} are illustrated from the top to bottom rows, \textit{i.e.}, ``\textit{cabinet, bed}'' (first and second rows), ``\textit{sofa, window}'' (third and forth rows), ``\textit{chair, floor}'' (fifth row), ``\textit{sink, toilet}'' (last row).}
	\label{fig:scannet-visualization}
\end{figure}

Figure \ref{fig:s3dis-visualization} and \ref{fig:scannet-visualization} show the qualitative results of our proposed method for 2-way 1-shot point cloud semantic segmentation on the S3DID and ScanNet dataset, respectively. We compare the predictions of one query point cloud from our method with the ground truths and predictions from ProtoNet.
As we can see from Figure \ref{fig:s3dis-visualization}, the S3DIS dataset is very challenging in many scenarios, \textit{e.g.}, ``the white columns that are very similar to the white wall and the window frame'' (first row of Figure \ref{fig:s3dis-visualization}), ``the doors that only have visible door frames'' (second row of Figure \ref{fig:s3dis-visualization}), ``the table that has a lot of clutter on it'' (last row of Figure \ref{fig:s3dis-visualization}). 
The accuracy of the predictions from our method drops for these challenging scenarios due to the limitation of having only one labeled sample as support. Nonetheless, our method still generally gives more accurate segmentation results than ProtoNet in all cases (\eg our method nicely segments the `ceiling', `floor', `chair', `bookcase', `table' in each scene from top to bottom in Figure \ref{fig:s3dis-visualization}).

In contrast to the S3DIS dataset, the ScanNet dataset contains more diverse room types, such as \textit{bathroom} (see first and last rows of Figure \ref{fig:scannet-visualization}),  \textit{bedroom/hotel} (see second row of Figure \ref{fig:scannet-visualization}), \textit{living room/lounge} (see third and fifth rows of Figure \ref{fig:scannet-visualization}), \textit{etc}. Our proposed method is able to correctly segment most of semantic classes within these new room types, while ProtoNet gives poor segmentation results that mix the background class with the semantic classes. We believe that our correct segmentations are consequences of integrating the attention-aware feature embedding and multi-prototype transductive inference, which facilitates the smoothness among neighboring points and the distinction between different semantic classes.

\section{Conclusion}
This paper investigates the unexplored yet important few-shot point cloud semantic segmentation problem. We propose a novel solution: the attention-aware multi-prototype transductive inference method, which achieves clear and consistent improvements over baselines on a variety of few-shot point cloud semantic segmentation tasks. Furthermore, this work offers several key insights on few-shot 3D point cloud semantic segmentation. Firstly, the learning of the discriminative features that encode both geometric and semantic context is the core of the metric-based few-shot point cloud semantic segmentation method. Secondly, the data distributions of 3D point clouds are complex and cannot be sufficiently modeled by a uni-modal distribution. Thirdly, the exploitation of intrinsic relationships in the embedding space is necessary for the point cloud segmentation task. 
Future work could investigate an adaptive generation of multi-prototypes to efficiently infer the number of prototypes based on data complexity. 

\section{Acknowledgements}
This research is supported in part by the National Research Foundation, Singapore under its IRC@SG Funding Initiative and partially performed on resources of the National Supercomputing Centre, Singapore. It is also partially supported by the Tier 2 grant MOE-T2EP20120-0011 from the Singapore Ministry of Education.

%\newpage
\appendix
\section{Supplementary Material}
This supplementary contains the splitting details of the S3DIS and ScanNet datasets (Section \ref{sec:dataset_split}), more framework details including the architecture of EdgeConv (Section \ref{sec:edgeconv}) and the settings of three hyper-parameters (Section \ref{sec:hyperparams}).

\subsection{Dataset Split}\label{sec:dataset_split}
Table \ref{tbl:data_split} lists the class names in each split of the S3DIS and ScanNet datasets.

\begin{table}[H]
	\centering
	\scalebox{0.8}{
		\begin{tabular}{p{1.2cm} | m{3.5cm} | m{3.5cm} }\hline \toprule[0.6pt]
			&~~~~~~~~~~~~~~split=0 
			&~~~~~~~~~~~~~~split=1 \\\hline
			\textbf{S3DIS} & beam, board, bookcase, ceiling, chair, column
			& door, floor, sofa, table, wall, window~ \\\hline
			\textbf{ScanNet} & bathtub, bed, bookshelf, cabinet, chair, counter, curtain, desk, door, floor & otherfurniture, picture, refrigerator, show curtain, sink, sofa, table, toilet, wall, window~ \\\hline
			\toprule[0.6pt]
	\end{tabular}}
	\caption{Test class names for each split of S3DIS and ScanNet.}
	\label{tbl:data_split}
\end{table}

\subsection{More Framework Details}
\subsubsection{EdgeConv architecture details}\label{sec:edgeconv}
\begin{figure}[h!]
	\centering
	\includegraphics[scale=0.55]{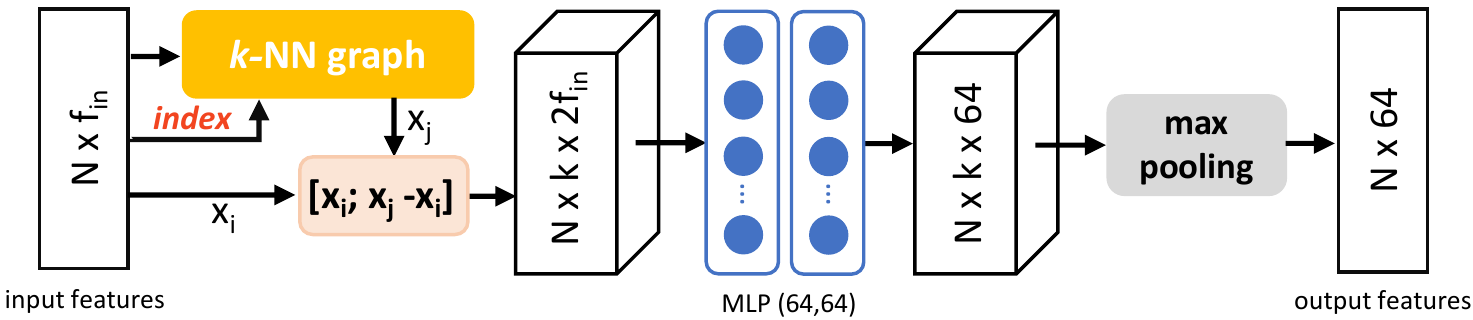}
	\caption{The architecture of EdgeConv component in the embedding network.}
	\label{fig:edgeconv}
\end{figure}

Figure \ref{fig:edgeconv} illustrates the architecture and configuration of EdgeConv, which is a basic block of the feature extractor. To perform graph CNN, a $k$-NN graph is dynamically computed from the input point-wise features to EdgeConv. 
Note that this $k$-NN graph is different from the $k$-NN graph in Section 3.2.3.
We set $k=20$ in our experiments. Each point $\textbf{x}_i$ in the point cloud is concatenated with its translated neighbor point $(\textbf{x}_j-\textbf{x}_i)$, which is yielded by translating $\textbf{x}_j$ to the local system with $\textbf{x}_i$ as the center. Consequently, a $N\times k \times 2f_{in}$ feature tensor is produced from the input tensor $N\times f_{in}$ and further passed to two MLP layers. Finally, EdgeConv aggregates the resultant feature tensor over the $k$ neighboring features using a max-pooling operator to generate the output point-wise features.

\subsubsection{Hyper-parameter settings}\label{sec:hyperparams}
As mentioned in Section \ref{sec:results}, we empirically find that the optimal value of $\sigma$ varies in different datasets. Additionally, we also observe varying optimal number of prototypes per class $n$ under different few-shot settings. Table \ref{tbl:n_setting} shows the optimal value of $n$ in different few-shot settings. It can be seen that $n$ becomes larger when the number of shots increases. This is reasonable since more shots result in larger number of observed support points for each class, which requires larger $n$ to model the  larger variety. From Table \ref{tbl:n_setting}, we also observe that $n$ becomes larger when the number of ``ways" increases. This is probably due to the more difficult 3-way segmentation requires fine-grained multi-prototypes for each class. We set $k=200$ for the $k$-NN graph mentioned in Section \ref{sec:transductive_inference} on all few-shot settings in both datasets.

\begin{table}[H]
	\centering
	\scalebox{0.82}{
		\begin{tabular}{c|c c c c}\hline\toprule[0.6pt]
			& 2-way 1-shot & 2-way 5-shot & 3-way 1-shot & 3-way 5-shot \\\hline
			$n=$  & 100 & 150 & 150 & 300 \\\hline\toprule[0.6pt]
	\end{tabular}}
	\caption{The value of $n$ in different few-shot settings.}
	\label{tbl:n_setting}
\end{table}

{\small
	\bibliographystyle{ieee_fullname}
	\bibliography{egbib}
}

\end{document}